\documentclass[11pt,a4paper]{article}
\usepackage[hyperref]{acl2021}
\usepackage{times}
\usepackage{latexsym}
\usepackage{caption}
\usepackage{url}
\usepackage[titletoc,title]{appendix}
\usepackage{soul}
\usepackage{tabularx}
\usepackage{graphicx}
\usepackage{caption}
\DeclareCaptionFont{small}{\small}
\usepackage{amsthm}
\theoremstyle{definition}

\theoremstyle{remark}

\usepackage{amssymb}
\usepackage{url}
\usepackage{makecell}

\RequirePackage{filecontents}
\usepackage[strict]{changepage}
\usepackage{array,multirow}
\usepackage{booktabs} 
\usepackage{CJKutf8}
\usepackage{makecell}
\usepackage{subfig}

\aclfinalcopy 
\def\aclpaperid{8} 

\title{
  Paraphrases as Foreign Languages \\in
  Multilingual Neural Machine Translation
}

\author{
  Zhong Zhou\\
  {\small Carnegie Mellon University}\\
  {\small \tt zhongzhou@cmu.edu} 
  \\\And 
  Matthias Sperber\\
  {\small Karlsruhe Institute of Technology}\\
  {\small \tt matthias.sperber@kit.edu}
  \\\And 
  Alex Waibel\\
  \small Carnegie Mellon University\\
  \small Karlsruhe Institute of Technology\\
  {\small \tt alex@waibel.com }
}

\date{}

\begin{document}

\maketitle

\begin{abstract}
  Paraphrases, the rewordings of the same semantic meaning, 
  are useful for improving generalization and 
  translation.
  However, prior works only explore
  paraphrases at the word or phrase level
  , not at the sentence or corpus level. 
  Unlike previous works that only explore paraphrases
  at the word or phrase level, we use different 
  translations of the whole training data that are 
  consistent in structure as paraphrases at the 
  corpus level.
  We train on parallel 
  paraphrases in multiple languages from various sources. 
  We treat paraphrases as foreign languages, 
  tag source sentences with paraphrase 
  labels, and train on parallel paraphrases
  in the style of multilingual 
  Neural Machine Translation (NMT).
  Our multi-paraphrase NMT
  that trains only on two languages
  outperforms the multilingual baselines.
  Adding paraphrases improves the rare word translation and 
  increases entropy and diversity in lexical
  choice. Adding the source 
  paraphrases boosts performance better than adding 
  the target ones.
  Combining both the source 
  and the target paraphrases lifts performance 
  further; combining paraphrases with 
  multilingual data helps but has mixed 
  performance. 
  We achieve a BLEU score
  of 57.2 for French-to-English translation using 24 corpus-level 
  paraphrases of the Bible, which
  outperforms the multilingual baselines
  and is +34.7 above 
  the single-source single-target NMT baseline.
\end{abstract}

\begin{figure}[!t]
  \small
  \hspace*{-0.6cm} 
  \subfloat[][multilingual NMT]{
    \includegraphics[width=3.5in]{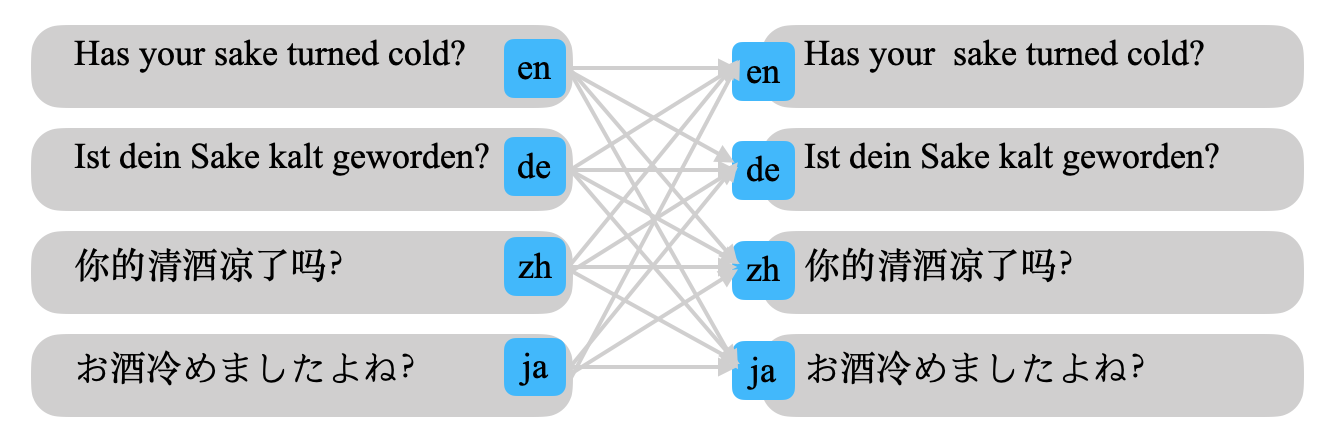}
  }\\
  \hspace*{-0.6cm}
  \subfloat[][multi-paraphrase NMT]{
    \includegraphics[width=3.5in]{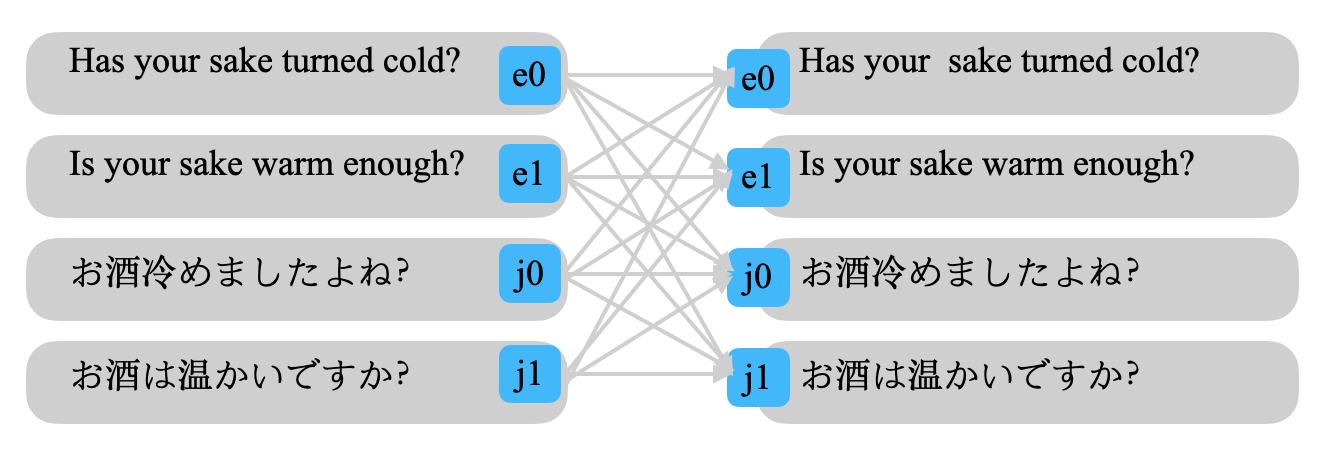}
  }
  \caption{Translation Paths in (a) multilingual NMT (b) multi-paraphrase NMT. Both form almost a complete bipartite graph. }
  \label{fig:new_graph}
\end{figure}

\section{Introduction} \label{introduction}
Paraphrases, rewordings of texts with  
preserved semantics, are often used to 
improve generalization and the sparsity issue in translation
\cite{callison2006improved, fader2013paraphrase, ganitkevitch2013ppdb, narayan2017split, sekizawa2017improving}.
Unlike previous works that use paraphrases at the word/phrase 
level, we research on different translations of the 
whole corpus that are consistent in structure 
as paraphrases at the corpus level; we refer to paraphrases
as the different translation versions of the
same corpus. We train paraphrases in the style of multilingual NMT
\cite{johnson2017google, ha2016toward}
.
Implicit parameter sharing 
enables multilingual NMT to learn across 
languages and achieve better generalization
\cite{johnson2017google}. 
Training on closely 
related languages are shown to
improve translation 
\cite{zhong2018massively}. We view 
paraphrases as an extreme case of closely related 
languages and view multilingual data as paraphrases 
in different languages. Paraphrases can 
differ randomly or systematically as each carries 
the translator's unique style. 

We treat paraphrases as foreign languages,
and train a unified NMT model on paraphrase-labeled data
with a shared attention in the style of multilingual NMT.
Similar to multilingual NMT's objective of translating
from any of the $N$ input languages to any of the
$M$ output languages \cite {firat2016multi}, multi-paraphrase 
NMT aims to translate from any of the $N$ input paraphrases to any of the
$M$ output paraphrases in Figure~\ref{fig:new_graph}. In Figure~\ref{fig:new_graph}, we see different expressions of a host
showing courtesy to a guest to ask whether sake (a type of alcohol drink that is normally served warm in Asia)
needs to be warmed.  In Table~\ref{table:bible}, we show a few examples of parallel paraphrasing data in the Bible corpus. Different translators' styles
give rise to rich parallel paraphrasing
data, covering wide range of domains. In Table~\ref{table:if}, we also show 
some paraphrasing examples from the modern poetry dataset, which we are considering for future research. 

Indeed,
we go
beyond the traditional NMT learning of one-to-one
mapping between the source and the target text;
instead,
we exploit the many-to-many mappings between the source
and target text through training on paraphrases that are consistent
to each other at the corpus level.
Our method achieves high translation performance and gives
interesting findings. 
The differences between our work and the prior works are mainly the following.

Unlike previous works that use paraphrases at the word or phrase level,
we use paraphrases at the entire corpus level to improve
translation performance. We use different translations of the whole training
data consistent in structure as paraphrases of the
full training data. Unlike most of the multilingual NMT works that uses data from
multiple languages, we use paraphrases as foreign languages in
a single-source single-target NMT system training only on data from the source
and the target languages. 

%

Our main findings in harnessing paraphrases in NMT are the following. 
\begin{enumerate}
\item{Our multi-paraphrase NMT results
  show significant improvements in BLEU scores
  over all baselines. }
\item{Our paraphrase-exploiting NMT uses only two languages, the source
  and the target languages, and achieves higher BLEUs than the multi-source and
  multi-target NMT that incorporates more languages. }
\item{We find that adding the source paraphrases helps better
  than adding the target paraphrases.}
\item{We find that adding paraphrases at both the source and the target
  sides is better than adding at either side. } 
\item{We also find that adding
  paraphrases with additional multilingual data yields mixed performance; its performance is better than training on language families alone, but is worse than training on both the source and target paraphrases without language families.} 
\item{Adding paraphrases improves
  the sparsity issue of rare word translation
  and diversity in lexical choice.}
\end{enumerate}
%

In this paper, we begin with introduction and
related work in Section
\ref{introduction} and \ref{relatedwork}.
We introduce our models
in Section \ref{methodology}.
Finally, we present our results
in Section \ref{experiments} and
conclude in Section \ref{conclusion}.

\section{Related Work} \label{relatedwork}
\begin{figure}[t]
  \small
  \centering
  \includegraphics[width=3.1in]{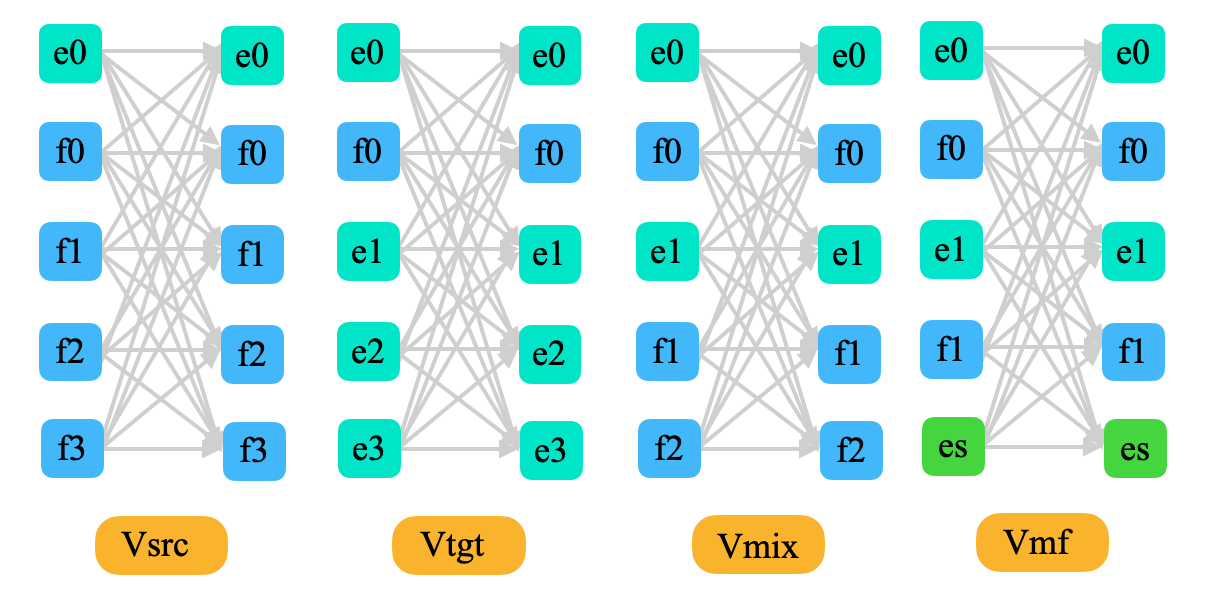}
  \caption{Examples of different ways of adding 5 paraphrases. \texttt{e[?n]} and \texttt{f[?n]} refers to different English and French paraphrases, \texttt{es} refers to the Spanish (an example member of Romance family) data. We always evaluate the translation path from \texttt{f0} to \texttt{e0}. }
\end{figure}
\subsection{Paraphrasing}
Many works generate and harness paraphrases
\cite{barzilay2001extracting, pang2003syntax, callison2005scaling, mallinson2017paraphrasing, ganitkevitch2013ppdb, brad2017neural, quirk2004monolingual, madnani2012re, suzuki2017building, hasan2016neural}.
Some are on question and answer 
\cite{fader2013paraphrase, dong2017learning}, evaluation of translation \cite{zhou2006re} 
and more recently NMT 
\cite{narayan2017split, sekizawa2017improving}.
Past research includes
paraphrasing unknown
words/phrases/sub-sentences \cite{callison2006improved, narayan2017split, sekizawa2017improving, fadaee2017data}.
These approaches are similar in transforming the 
difficult sparsity problem of rare words prediction and long sentence
translation into a simpler problem with known words and
short sentence translation.
It is worthwhile to contrast paraphrasing that diversifies
data, with knowledge distillation that benefits from making data
more consistent \cite{gu2017non}.

Our work is different
in that we exploit paraphrases at the corpus
level, rather than at the word or phrase level. 
\subsection{Multilingual Attentional NMT}
Machine polyglotism which trains 
machines to translate  
any of the $N$ input languages to any of the
$M$ output languages
is a new paradigm in multilingual NMT
\cite{firat2016multi, zoph2016multi, dong2015multi, gillick2016multilingual, al2013polyglot, tsvetkov2016polyglot}. 
Many multilingual NMT systems involve multiple encoders
and decoders \cite{ha2016toward}, and it is hard to combine
attention for quadratic language pairs bypassing
quadratic attention mechanisms
\cite{firat2016multi}. An interesting work is training 
a universal model with a shared attention mechanism
with the source and target language labels and
Byte-Pair Encoding (BPE)
\cite{johnson2017google, ha2016toward}.
This method is elegant in its 
simplicity and its 
advancement in
low-resource language 
translation and zero-shot 
translation using
pivot-based translation mechanism 
\cite{johnson2017google, firat2016multi}. 

Unlike previous works, our parallelism
is across paraphrases, not across
languages.
In other words, we achieve higher translation
performance in the single-source single-target paraphrase-exploiting
NMT than that of the multilingual NMT.

\begin{table}[t]
  \hspace*{-0.2cm}
  \small
  \centering
  \begin{tabularx}{\columnwidth}{|X|X|X|X|X|} \hline
    Data & 1 & 6 & 11 & 13  \\ \hline \hline
    \textit{\texttt{Vsrc}} & 22.5 & 41.4 & 48.9 & 48.8 \\ \hline
    \textit{\texttt{Vtgt}} & 22.5 & 40.5 & 47.0 & 47.4 \\ \hline
  \end{tabularx}
  \caption{Comparison of adding source paraphrases and adding target paraphrases.
    All acronyms including data are explained in Section \ref{experimentsbase}.
  }
  \label{table:srcVStarget}
\end{table}
\begin{table}[t]
  \hspace*{-0.2cm}
  \small
  \centering
  \begin{tabularx}{\columnwidth}{|p{1.15cm}|X|X|X|X|X|X|} \hline
    data & 1 & 6 & 11 & 16 & 22 & 24\\ \hline \hline
    \textit{\texttt{WMT}} & 22.5 & 30.8 & 29.8 & 30.8 & 29.3 & - \\ \hline
    \textit{\texttt{Family}} & 22.5 & 39.3 & 45.4 & 49.2 & 46.6 & - \\ \hline
    \textit{\texttt{Vmix}} & 22.5 & 44.8 & 50.8 & 53.3 & 55.4 & 57.2 \\ \hline
    \textit{\texttt{Vmf}} & - & - & 49.3 & - & - & - \\ \hline
  \end{tabularx}
  \caption{Comparison of adding a mix of the source paraphrases and the target paraphrases against the baselines. All acronyms including data are explained in Section \ref{experimentsbase}.
  }
  \label{table:bleu}
\end{table}
\section{Models} \label{methodology}
We have four baseline models. Two are single-source
single-target attentional NMT models, the other two are multilingual
NMT models with a shared attention
\cite{johnson2017google, ha2016toward}.
In Figure~\ref{fig:new_graph}, we show an example of multilingual attentional
NMT. Translating from all 4 languages to each other, we have 12 translation paths.
For each translation path, we
label the source sentence with the source and target language tags.
Translating from \begin{CJK*}{UTF8}{gbsn}``你的清酒凉了吗?''\end{CJK*}
to ``Has your sake turned cold?'', we label
the source sentence with \texttt{\_\_opt\_src\_zh \_\_opt\_tgt\_en}.
More details are in Section~\ref{experiments}.

In multi-paraphrase model, all source sentences
are labeled with the paraphrase tags.
For example, in French-to-English translation,
a source sentence may be tagged with \texttt{\_\_opt\_src\_f1 \_\_opt\_tgt\_e0},
denoting that it is translating from version ``f1'' of French data to
version ``e0'' of English data. 
In Figure~\ref{fig:new_graph},
we show 2 Japanese and 2 English paraphrases.
Translating from all 4
paraphrases to each other ($N=M=4$),
we have 12 translation paths as $N\times(N-1)=12$.
For each translation path, we
label the source sentence with the source and target
paraphrase tags. For
the translation path from
\begin{CJK}{UTF8}{min}``お酒冷めましたよね?''\end{CJK}
to ``Has your sake turned cold?'', we label
the source sentence with \texttt{\_\_opt\_src\_j1 \_\_opt\_tgt\_e0} in Figure~\ref{fig:new_graph}.
Paraphrases of the same translation path carry the same labels.
Our paraphrasing data is at the corpus level, and
we train a unified NMT model with a shared attention.
Unlike the
paraphrasing sentences in Figure~\ref{fig:new_graph},
We show this example
with only one sentence, it is similar when the training data contains many sentences.  
All sentences in the same paraphrase path share the same labels.  

  \begin{table*}[]
  \small
  \centering
  \begin{adjustwidth}{-0.37cm}{-0.37cm}
    \begin{tabularx}{1.08\textwidth}{|X|X|X|} \hline
    Source Sentence
    & Machine Translation
    & Correct Target Translation\\ \hline \hline
    Comme de l'eau fra{\^i}che pour une personne fatigu{\'e}, Ainsi est une bonne nouvelle venant d'une terre lointaine.
      & As cold waters to a thirsty soul, so is good news from a distant land.
      & Like cold waters to a weary soul, so is a good report from a far country.
    \\ \hline
        Lorsque tu seras invit{\'e} par quelqu'un {\`a} des noces, ne te mets pas {\`a} la premi{\`e}re place, de peur qu'il n'y ait parmi les invit{\'e}s une personne plus consid{\'e}rable que toi,
      & When you are invited to one to the wedding, do not be to the first place, lest any one be called greater than you.
      & When you are invited by anyone to wedding feasts, do not recline at the chief seat lest one more honorable than you be invited by him,
    \\ \hline
    Car chaque arbre se conna{\^i}t {\`a} son fruit. On ne cueille pas des figues sur des {\'e}pines, et l'on ne vendange pas des raisins sur des ronces.
      & For each tree is known by its own fruit. For from thorns they do not gather figs, nor do they gather grapes from a bramble bush.
      & For each tree is known from its own fruit. For they do not gather figs from thorns, nor do they gather grapes from a bramble bush.
    \\ \hline
    Vous tous qui avez soif, venez aux eaux, Même celui qui n'a pas d'argent! Venez, achetez et mangez, Venez, achetez du vin et du lait, sans argent, sans rien payer!
    &Come, all you thirsty ones, come to the waters; come, buy and eat. Come, buy for wine, and for nothing, for without money.
    & Ho, everyone who thirsts, come to the water; and he who has no silver, come buy grain and eat. Yes, come buy grain, wine and milk without silver and with no price.
    \\ \hline
    Oui , vous sortirez avec joie , Et vous serez conduits en paix ; Les montagnes et les collines éclateront d'allégresse devant vous , Et tous les arbres de la campagne battront des mains .
    & When you go out with joy , you shall go in peace ; the mountains shall rejoice before you , and the trees of the field shall strike all the trees of the field .
    & For you shall go out with joy and be led out with peace . The mountains and the hills shall break out into song before you , and all the trees of the field shall clap the palm .
    \\ \hline
  \end{tabularx}
  \end{adjustwidth}
  \caption{Examples of French-to-English translation trained using 12 French paraphrases and 12 English paraphrases.}
  \label{table:french}
\end{table*}
\section{Experiments and Results}\label{experiments}
\subsection{Data}\label{experimentsdata}
Our main data is the French-to-English Bible corpus
\cite{mayer2014creating}, containing 12 versions
of the English Bible and 12 versions of the French Bible
\footnote{We considered the open subtitles
  with different scripts of
  the same movie in the same language; they
  covers many topics, but they are noisy and
  only differ in interjections.
  We also considered the poetry dataset where a poem
  like ``If'' by Rudyard Kipling
  is translated many times, 
  by various people into the same language,
  but the data is small.}.
We translate from French to English. 
Since these 24 translation versions are consistent in structure,
we refer to them as paraphrases at corpus level.
In our paper, each paraphrase refers to
each translation version of whole Bible corpus.
To understand our setup, if we use all
12 French paraphrases and all 12 English paraphrases
so there are 24 paraphrases in total, i.e., $N=M=24$,
we have 552 translation paths because $N\times(N-1)=552$.
The original corpus contains missing or extra verses for different
paraphrases; we clean and align 24
paraphrases of the Bible corpus and randomly sample the training, validation and test
sets according to the 0.75, 0.15, 0.10 ratio.
Our training set contains only 23K verses,
but is massively parallel across paraphrases.

For all experiments, we choose a specific English corpus
as \texttt{e0} and a specific French corpus as \texttt{f0} which we evaluate across all
experiments to ensure consistency in comparison, and 
we evaluate all translation performance from \texttt{f0} to \texttt{e0}.

\subsection{Training Parameters}\label{experimentspara}
  In all our experiments,
we use a minibatch size of 64,
dropout rate of 0.3,
4 RNN layers of size 1000,
a word vector size of 600,
a learning rate of 0.8 that decays at
the rate of 0.7 if the validation score
is not improving
or it is past epoch 9
across all LSTM-based experiments. We train 13 epochs.
Byte-Pair Encoding (BPE) is used at preprocessing stage \cite{ha2016toward}.
Our code is built on OpenNMT \cite{klein2017opennmt} and we evaluate our models using
BLEU scores \cite{papineni2002bleu}, entropy \cite{shannon1951prediction},
F-measure and qualitative evaluation. 

\subsection{Baselines}\label{experimentsbase}
\begin{figure}[t]
  \small
  \centering
  \hspace*{-0.2cm}
  \includegraphics[width=3.6in]{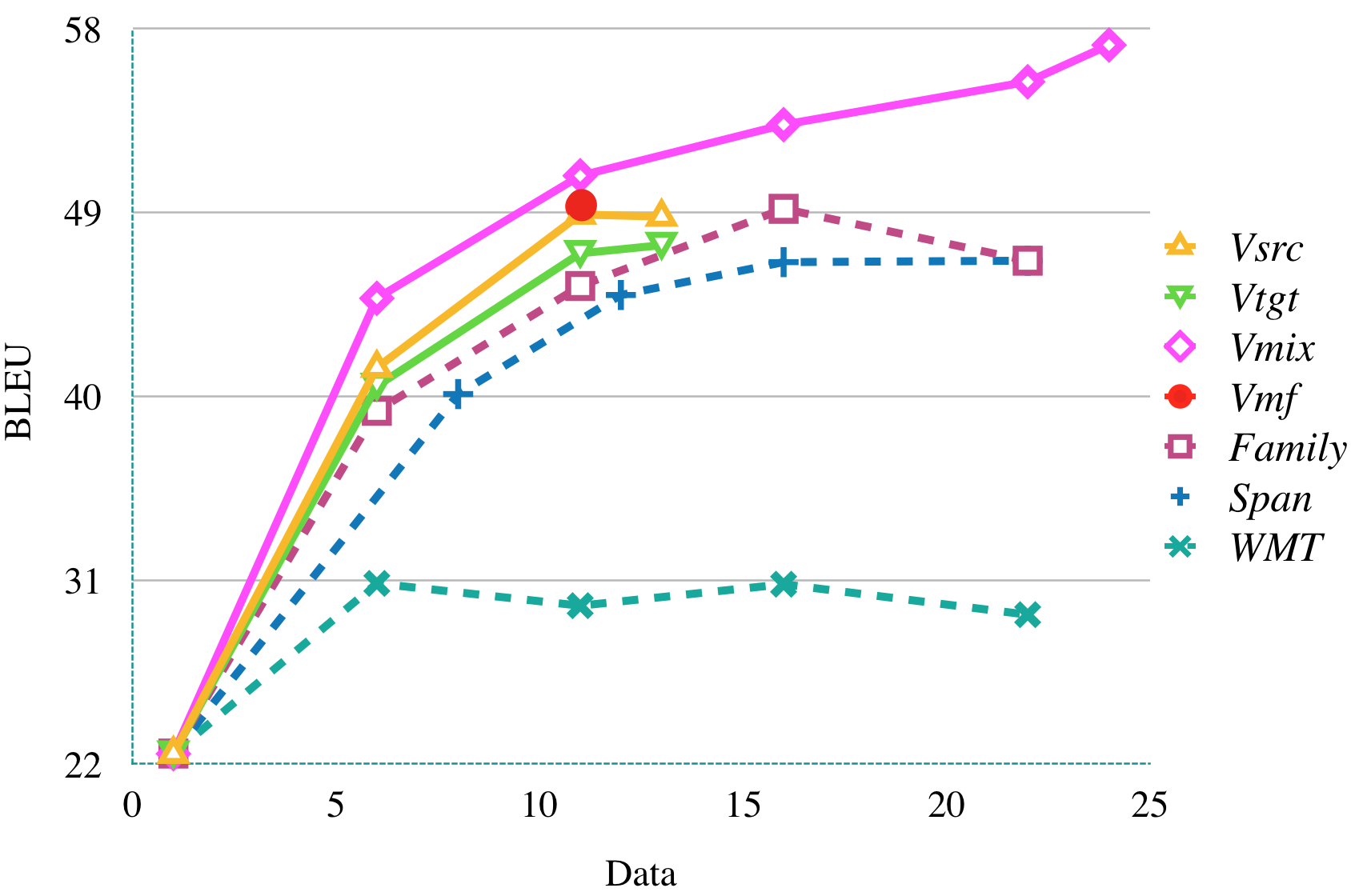}
  \caption{BLEU plots showing the effects of different ways of adding training data in French-to-English Translation. All acronyms including data are explained in Section \ref{experimentsbase}. 
    }
  \label{fig:multi_family1}
\end{figure}

We introduce a few acronyms for our four baselines to describe the experiments
in Table~\ref{table:srcVStarget}, Table~\ref{table:bleu}
and Figure~\ref{fig:multi_family1}. Firstly, we have two single-source
single-target attentional NMT models, \textit{\texttt{Single}} and
\textit{\texttt{WMT}}. \textit{\texttt{Single}} trains
on \texttt{f0} and \texttt{e0} and gives a BLEU of 22.5,
the starting point for all curves in Figure~\ref{fig:multi_family1}.
\textit{\texttt{WMT}} adds the
out-domain WMT'14 French-to-English data
on top of \texttt{f0} and \texttt{e0}; it serves as
a weak baseline that helps us to evaluate all
experiments' performance
discounting the effect of increasing data.

Moreover, we have two multilingual baselines\footnote{
  For multilingual baselines, we use the
  additional Bible corpus in
  22 European languages that are cleaned and
  aligned to each other.} built on multilingual
attentional NMT, \textit{\texttt{Family}}
and \textit{\texttt{Span}} \cite{zhong2018massively}. 
\textit{\texttt{Family}} refers to the multilingual
baseline by adding one language family at a time,
where on top of the French corpus \texttt{f0}
and the English corpus \texttt{e0}, we add up to 20
other European languages.
\textit{\texttt{Span}} refers
to the multilingual
baseline by adding one \textit{span} at a time,
where a span is a set of languages
that contains at least one language
from all the families in the data; in
other words, span is a sparse
representation of all the families.
Both \textit{\texttt{Family}}
and \textit{\texttt{Span}}
trains on the Bible in 22 Europeans languages trained using
multilingual NMT.
Since \textit{\texttt{Span}} is always suboptimal
to \textit{\texttt{Family}} in our results,
we only show numerical results
for \textit{\texttt{Family}} in Table~\ref{table:srcVStarget}
and \ref{table:bleu}, and we plot both \textit{\texttt{Family}}
and \textit{\texttt{Span}} in Figure~\ref{fig:multi_family1}.
The two multilingual baselines are strong baselines while
the f\textit{\texttt{WMT}} baseline is a weak baseline that helps us to evaluate all
experiments' performance
discounting the effect of increasing data.
All baseline results are taken
from a research work which uses the grid of
(1, 6, 11, 16, 22) for the number
of languages or equivalent number of
unique sentences and we follow the same in Figure~\ref{fig:multi_family1}
\cite{zhong2018massively}.
All experiments for each grid point carry the same number of unique sentences.

\begin{table}[t]
  \small
  \hspace*{-0.2cm}
  \centering
  \begin{tabular}{ | p{1.15cm} | p{.9cm} | p{.9cm} | p{.9cm} | p{.9cm} | p{.9cm} | }\hline
    data & 6 & 11 & 16 & 22 & 24\\ \hline \hline
    Entropy & 5.6569 & 5.6973 & 5.6980 & 5.7341 & 5.7130\\ \hline
    Bootstrap 95\% CI &(5.6564, 5.6574)&(5.6967, 5.6979)&(5.6975, 5.6986)&(5.7336, 5.7346)&(5.7125, 5.7135)\\ \hline
    \textit{\texttt{WMT}} & - & 5.7412 & 5.5746 & 5.6351 & - \\ \hline
  \end{tabular}
  \caption{Entropy increases with the number of paraphrase corpora in \textit{\texttt{Vmix}}. The 95\% confidence interval is calculated via bootstrap resampling with replacement.
  }
  \label{table:entropy}
\end{table}
\begin{table}[t]
  \hspace*{-0.2cm}
  \small
  \centering
  \begin{tabular}{ | p{1.15cm} | p{.9cm} | p{.9cm} | p{.9cm} | p{.9cm} | p{.9cm} | }\hline
    data & 6 & 11 & 16 & 22 & 24\\ \hline \hline
    F1(freq1) & 0.43 & 0.54 & 0.57 & 0.58 & 0.62 \\ \hline
    \textit{\texttt{WMT}} & - & 0.00 & 0.01 & 0.01 & -  \\ \hline
  \end{tabular}
  \caption{F1 score of frequency 1 bucket increases with the number of paraphrase corpora in \textit{\texttt{Vmix}}, showing training on paraphrases improves the sparsity at tail and the rare word problem.}
  \label{table:f1}
\end{table}
Furthermore, \textit{\texttt{Vsrc}}
refers to adding more source (English) paraphrases,
and \textit{\texttt{Vtgt}} refers to
adding more target (French) paraphrases.
\textit{\texttt{Vmix}} refers to
adding both the source and the target paraphrases.
\textit{\texttt{Vmf}} refers to combining
\textit{\texttt{Vmix}} with additional multilingual data;
note that only \textit{\texttt{Vmf}},
\textit{\texttt{Family}} and \textit{\texttt{Span}} 
use languages other than French and English,
all other experiments use only English and French. For the x-axis,
data refers to the number of paraphrase corpora for
\textit{\texttt{Vsrc}}, \textit{\texttt{Vtgt}}, \textit{\texttt{Vmix}}; data refers
to the number of languages for \textit{\texttt{Family}}; data refers to
and the equivalent number of unique training sentences compared to other training curves for
\textit{\texttt{WMT}} and \textit{\texttt{Vmf}}.

\subsection{Results}
\textbf{\ul{Training on paraphrases gives better performance than all baselines:}} The
translation performance of training
on 22 paraphrases, i.e., 11 English paraphrases and 11 French paraphrases, 
achieves a BLEU	score of 55.4, which is +32.9 above the \textit{\texttt{Single}} baseline,
+8.8 above the \textit{\texttt{Family}} baseline,
and +26.1 above the \textit{\texttt{WMT}} baseline. 
Note that the \textit{\texttt{Family}} baseline
uses the grid of (1, 6, 11, 16, 22) for number of languages,
we continue to use this grid for our results on number of
paraphrases,
which explains why we pick 22 as an example here. 
The highest BLEU 57.2 is achieved when we train on 24
paraphrases, i.e., 12 English paraphrases
and 12 French paraphrases. 

\textbf{\ul{Adding the source paraphrases boosts translation performance more than adding the target paraphrases:}} The
translation performance	of adding
the source paraphrases is higher than that of adding
the target paraphrases. Adding the source
paraphrases diversifies the data,
exposes the model to more rare words,
and enables better generalization. 
Take the experiments training on 13 paraphrases for example,
training on the source (i.e., 12 French paraphrases and the English paraphrase \texttt{e0}) gives a BLEU score of 48.8,
which has a gain of +1.4 over 47.4, the
BLEU score of training on the target (i.e., 12 English paraphrases and the
French paraphrase \texttt{f0}).
This suggests that adding
the source paraphrases is more effective than adding
the target paraphrases.

\textbf{\ul{Adding paraphrases from both sides is better than adding paraphrases from either side:}}
The curve of adding paraphrases from both the source and the target sides
is higher than both the
curve of adding the target paraphrases and
the curve of adding the source paraphrases.
Training on 11 paraphrases from
both sides, i.e., a total of 22 paraphrases achieves a BLEU score of 50.8,
which is +3.8 higher than that of
training on the target
side only and +1.9 higher than that of
training on the source side only. 
The advantage of combining both sides is that we can
combine paraphrases from both the source and
the target to reach 24 paraphrases in total to achieve a BLEU score
of 57.2.

\textbf{\ul{Adding both paraphrases and language families yields mixed performance:}}
We conduct one more experiment combining the source and target paraphrases together with
additional multilingual data. This is the only experiment on paraphrases 
where we use multilingual data
other than only French and English data. The BLEU score is 49.3, 
higher than training on families alone, in fact,
it is higher than training on eight European families altogether. 
However, it is lower than training on English and French paraphrases alone.
Indeed, adding
paraphrases as foreign languages is effective,
however, when there is a lack of
data, mixing the paraphrases with
multilingual data is helpful. 

\textbf{\ul{Adding paraphrases increases entropy and diversity in lexical choice, and improves the
    sparsity issue of rare words: }} We
use bootstrap resampling and construct 95\% confidence intervals for entropies \cite{shannon1951prediction}
of all models of \textit{\texttt{Vmix}}, i.e., models adding
paraphrases at both the source and the target sides. We find that
the more paraphrases, the higher
the entropy, the more diversity
in lexical choice as shown in Table \ref{table:entropy}.
From the word F-measure shown in Table \ref{table:f1},
we find that the more paraphrases, the better the model handles the
sparsity of rare words issue. 
Adding paraphrases not only achieves much higher
BLEU score than the \textit{\texttt{WMT}} baseline, but
also handles the sparsity issue
much better than the \textit{\texttt{WMT}} baseline. 

\textbf{\ul{Adding paraphrases helps rhetoric translation and increases expressiveness:}} Qualitative
evaluation shows many cases where rhetoric translation is
improved by training on diverse sets of paraphrases.
In Table \ref{table:french},
Paraphrases help NMT to
use a more contemporary synonym of ``silver'', ``money'',
which is more direct and
easier to understand.
Paraphrases simplifies the rhetorical or subtle
expressions, for example, our model uses ``rejoice'' to replace
``break out into song'', a personification device of mountains to describe joy, 
which captures the essence of the meaning being conveyed. 
However, we also observe
that NMT wrongly translates ``clap the palm'' to ``strike''.
We find the quality of rhetorical translation ties closely with
the diversity of parallel paraphrases data. 
Indeed, the use of paraphrases to improve rhetoric translation is
a good future research question.
Please refer to the Table \ref{table:french} for more qualitative examples.

\section{Conclusion} \label{conclusion}
We train on paraphrases as
foreign languages in the style of
multilingual NMT. Adding
paraphrases improves translation quality, the rare word issue,
and diversity in lexical choice. 
Adding the source paraphrases helps more than adding the target
ones, while combining both boosts performance further.
Adding multilingual data to paraphrases yields mixed performance. 
We would like to explore the common structure
and terminology consistency 
across paraphrases. As structure and terminology 
are shared across paraphrases, we are interested in building 
an explicit representation of the paraphrases and extending our work 
for better translation, or translation with more explicit and more 
explainable hidden states, which is very important in all neural 
systems. 

We are interested in broadening our dataset in our future experiments. 
We hope to use other parallel
paraphrasing corpora like the poetry
dataset as shown in Table~\ref{table:if}. There are very few poems that are
translated multiple times into the same
language, we therefore need to train
on extremely small dataset.
Rhetoric in paraphrasing is important in poetry
dataset, which again depends on the training
paraphrases. The limited data issue is also
relevant to the low-resource setting.

We would like to effectively train on extremely small
low-resource paraphrasing data. As discussed above about the 
potential research poetry dataset, dataset with multiple paraphrases 
is typically small and yet valuable. If we can train using extremely 
small amount of data, especially in the low-resource scenario, we would 
exploit the power of multi-paraphrase NMT further. 

Cultural-aware paraphrasing and subtle expressions are
vital \cite{levin1998interlingua, larson1984meaning}. Rhetoric in paraphrasing is a 
very important too. 
In Figure~\ref{fig:new_graph}, ``is your sake warm enough?'' in Asian culture 
is an implicit way of
saying ``would you like me to warm the sake for you?''.
We would like to model the culture-specific
subtlety through multi-paraphrase training. 

\bibliography{acl2019}
\bibliographystyle{acl_natbib}

\clearpage
\begin{appendices}
  \label{sec:supplemental}
  \section{Supplemental Materials}
  We show a few examples of parallel paraphrasing data in the Bible corpus. We also show some paraphrasing examples from the modern poetry dataset, which we are considering for future research.

  \begin{table*}[]
    \small
    \centering
    \begin{tabular}{ | p{2.9cm} | p{11.8cm} | }\hline\hline
      \multirow{4}{*}{English Paraphrases}
      & Consider the lilies, how they grow: they neither toil nor spin, yet I tell you, even Solomon in all his glory was not arrayed like one of these. \textit{English Standard Version}.\\\cline{2-2}
      & Look how the wild flowers grow! They don't work hard to make their clothes. But I tell you Solomon with all his wealth wasn't as well clothed as one of these flowers. \textit{Contemporary English Version}.\\\cline{2-2}
      & Consider how the wild flowers grow. They do not labor or spin. Yet I tell you, not even Solomon in all his splendor was dressed like one of these.  \textit{New International Version}.
      \\\hline\hline
      \multirow{4}{*}{French Paraphrases}
      & Considérez les lis! Ils poussent sans se fatiguer à tisser des vêtements. Et pourtant, je vous l’assure, le roi Salomon lui-même, dans toute sa gloire, n’a jamais été aussi bien vêtu que l’un d’eux! \textit{La Bible du Semeur}.\\\cline{2-2}
      & Considérez comment croissent les lis: ils ne travaillent ni ne filent; cependant je vous dis que Salomon même, dans toute sa gloire, n'a pas été vêtu comme l'un d'eux. \textit{Louis Segond}. \\\cline{2-2}
      & Observez comment poussent les plus belles fleurs: elles ne travaillent pas et ne tissent pas; cependant je vous dis que Salomon lui-même, dans toute sa gloire, n'a pas eu d’aussi belles tenues que l'une d'elles. \textit{Segond 21}.
      \\\hline\hline
      \multirow{4}{*}{Tagalog Paraphrases}
      & Wariin ninyo ang mga lirio, kung paano silang nagsisilaki: hindi nangagpapagal, o nangagsusulid man; gayon ma'y sinasabi ko sa inyo, Kahit si Salomon man, sa buong kaluwalhatian niya, ay hindi nakapaggayak na gaya ng isa sa mga ito. \textit{Ang Biblia 1978}. \\\cline{2-2}
      & Isipin ninyo ang mga liryo kung papaano sila lumalaki. Hindi sila nagpapagal o nag-iikid. Gayunman, sinasabi ko sa inyo: Maging si Solomon, sa kaniyang buong kaluwalhatian ay hindi nadamitan ng tulad sa isa sa mga ito. \textit{Ang Salita ng Diyos}. \\\cline{2-2}
      & Tingnan ninyo ang mga bulaklak sa parang kung paano sila lumalago. Hindi sila nagtatrabaho ni humahabi man. Ngunit sinasabi ko sa inyo, kahit si Solomon sa kanyang karangyaan ay hindi nakapagdamit ng singganda ng isa sa mga bulaklak na ito. \textit{Magandang Balita Biblia}.
      \\\hline\hline
      \multirow{4}{*}{Spanish Paraphrases}
      &	Considerad los lirios, cómo crecen; no trabajan ni hilan; pero os digo que ni Salomón en toda su gloria se vistió como uno de éstos. \textit{La Biblia de las Américas}. \\\cline{2-2}
      & Fíjense cómo crecen los lirios. No trabajan ni hilan; sin embargo, les digo que ni siquiera Salomón, con todo su esplendor, se vestía como uno de ellos. \textit{Nueva Biblia al Día}. \\\cline{2-2}
      &	Aprendan de las flores del campo: no trabajan para hacerse sus vestidos y, sin embargo, les aseguro que ni el rey Salomón, con todas sus riquezas, se vistió tan bien como ellas. \textit{Traducción en lenguaje actual}. 
      \\\hline\hline
    \end{tabular}
    \caption{Examples of parallel paraphrasing data with English, French, Tagalog and Spanish paraphrases in Bible translation. }
    \label{table:bible}
  \end{table*}
  
  \begin{table*}[]
    \small
    \centering
    \begin{tabular}{ | p{2.9cm} | p{11.8cm} | }\hline\hline
      English Original &
      If you can fill the unforgiving minute
      with sixty seconds' worth of distance run,
      yours is the Earth and everything that's in it,
      and—which is more—you'll be a Man, my son!
      ``if'', \textit{Rudyard Kipling}. \\\hline\hline
      \multirow{4}{*}{German Translations}
      &
      Wenn du in unverzeihlicher Minute
      Sechzig Minuten lang verzeihen kannst:
      Dein ist die Welt—und alles was darin ist—
      Und was noch mehr ist—dann bist du ein Mensch!
      Translation by \textit{Anja Hauptmann}.  \\\cline{2-2}
      &
      Wenn du erfüllst die herzlose Minute
      Mit tiefstem Sinn, empfange deinen Lohn:
      Dein ist die Welt mit jedem Attribute,
      Und mehr noch: dann bist du ein Mensch, mein Sohn!
      Translation by \textit{Izzy Cartwell}.  \\\cline{2-2}
      &
      Füllst jede unerbittliche Minute
      Mit sechzig sinnvollen Sekunden an;
      Dein ist die Erde dann mit allem Gute,
      Und was noch mehr, mein Sohn: Du bist ein Mann!
      Translation by \textit{Lothar Sauer}. \\\hline\hline
      \multirow{4}{*}{Chinese Translations}
      &
      \begin{CJK*}{UTF8}{gkai}
        若胸有激雷, 而能面如平湖, 则山川丘壑, 天地万物皆与尔共, 吾儿终成人也！
      \end{CJK*} Translation by \textit{Anonymous}.
        \\\cline{2-2}
      &
        \begin{CJK*}{UTF8}{gkai}
        如果你能惜时如金利用每一分钟不可追回的光阴；那么，你的修为就会如天地般博大，并拥有了属于自己的世界，更重要的是：孩子，你成为了真正顶天立地之人！
        \end{CJK*} Translation by \textit{Anonymous}.
        \\\cline{2-2}
      &
          \begin{CJK*}{UTF8}{gkai}
            假如你能把每一分宝贵的光阴,
化作六十秒的奋斗—-你就拥有了整个世界, 最重要的是——你就成了一个真正的人，我的孩子！
          \end{CJK*} Translation by \textit{Shan Li}.
      \\ \hline  \hline
      \multirow{4}{*}{Portuguese Translations}
      &
      Se você puder preencher o valor do inclemente minuto perdido
      com os sessenta segundos ganhos numa longa corrida,
      sua será a Terra, junto com tudo que nela existe,
      e—mais importante—você será um Homem, meu filho!
      Translation by \textit{Dascomb Barddal}. \\\cline{2-2}
      &
      Pairando numa esfera acima deste plano,
      Sem receares jamais que os erros te retomem,
      Quando já nada houver em ti que seja humano,
      Alegra-te, meu filho, então serás um homem!...
      Translation by \textit{Féliz Bermudes}.\\\cline{2-2}
      &
      Se és capaz de dar, segundo por segundo,
      ao minuto fatal todo valor e brilho.
      Tua é a Terra com tudo o que existe no mundo,
      e—o que ainda é muito mais—és um Homem, meu filho!
      Translation by \textit{Guilherme de Almeida}. \\\hline\hline
    \end{tabular}
    \caption{Examples of parallel paraphrasing data with German, Chinese, and Portuguese paraphrases of the English poem ``If'' by Rudyard Kipling.}
    \label{table:if}
  \end{table*}
  \end{appendices}
\end{document}